\documentclass{llncs}
\pdfoutput=1
\usepackage{graphicx}
\usepackage[rightcaption]{sidecap}
\usepackage{cite}
\usepackage{amsmath,amsfonts}
\usepackage{array,booktabs}
\usepackage{url}
\usepackage{hyperref}

\newcommand{\RR}{\mathbb{R}}

\newcommand{\sign}{\text{sign}}
\providecommand{\B}[1]{\mathbf{#1}}
\def \lb {{\langle}}
\def \rb {{\rangle}}

\begin{document}
\mainmatter              
\title{Learning to rank from medical imaging data}
\titlerunning{Learning to rank from medical imaging data}
%
\author{Fabian Pedregosa\inst{1,2,4}
  \and Elodie Cauvet\inst{3,2}
  \and Ga\"el Varoquaux\inst{1,2}
  \and Christophe Pallier\inst{3,1,2}
  \and Bertrand Thirion\inst{1,2}
  \and Alexandre Gramfort\inst{1,2}}
\authorrunning{Fabian Pedregosa et al.} 
%
%
\institute{Parietal Team, INRIA Saclay-\^{I}le-de-France, Saclay, France\\
\email{fabian.pedregosa@inria.fr},
\and
CEA, DSV, I\textsuperscript{2}BM, Neurospin b\^{a}t 145,
91191 Gif-Sur-Yvette, France
\and
Inserm, U992, Neurospin b\^{a}t 145, 91191 Gif-Sur-Yvette, France
\and
SIERRA Team, INRIA Paris - Rocquencourt, Paris, France
}

\maketitle              

\begin{abstract}
  Medical images can be used to predict a clinical score coding for
  the severity of a disease, a pain level or the complexity of a
  cognitive task. In all these cases, the predicted variable has
  a natural order. While a standard classifier discards this information,
  we would like to take it into account in order to improve prediction
  performance. A standard linear regression does model such information,
  however the linearity assumption is likely not be satisfied
  when predicting from pixel intensities in an image.
  In this paper we address these modeling challenges
  with a supervised learning procedure where the model aims to order or rank images.
  We use a linear model for its robustness in high dimension and
  its possible interpretation.
  We show on simulations and two fMRI
  datasets that this approach is able to predict the correct ordering
  on pairs of images, yielding higher prediction accuracy than
  standard regression and multiclass classification techniques.
\end{abstract}

%
\section{Introduction}

%
Statistical machine learning has recently gained interest in the field
of medical image analysis. It is particularly useful for instance to
learn imaging biomarkers for computer aided diagnosis or as a way to
obtain additional therapeutic indications. These techniques are
particularly relevant in brain imaging \cite{cuingnet2011}, where the
complexity of the data renders visual inspection or univariate
statistics unreliable. A spectacular application of learning from
medical images is the prediction of behavior from functional MRI
activations~\cite{haynes2006}. Such a supervised learning problem can
be based on regression~\cite{kay2008} or classification
tasks~\cite{haxby2001}.

%
In a classification setting, \emph{e.g.} using support vector machines
(SVM)~\cite{LaConte2005317}, class labels are treated as an unordered
set. However, it is often the case that labels corresponding to
physical quantities can be naturally ordered: clinical scores, pain
levels, the intensity of a stimulus or the complexity of a cognitive
task are examples of such naturally ordered quantities. Because
classification models treat these as a set of classes, the
intrinsic order is ignored leading to suboptimal results. On the other
hand, in the case of linear regression models such as Lasso
\cite{Liu_Palatucci_Zhang_2009} or Elastic Net
\cite{Carroll_Cecchi_Rish_Garg_Rao_2009}, the explained variable is
obtained by a linear combination of the variables, the pixel
intensities in the present case, which benefits to the
interpretability of the model~\cite{Carroll_Cecchi_Rish_Garg_Rao_2009,
  Michel2011}. The limitation of regression models is that they assume
linear relationship between the data and the predicted variable. This
assumption is a strong limitation in practical cases. For instance, in
stroke studies, light disabilities are not well captured by the
standard NIHSS score. For this reason most studies use a
classification approach, forgoing the quantitative assessment of
stroke severity and splitting patients in a small number of different
classes. A challenge is therefore to work with a model that is able
to learn a non-linear relationship between the data and the target
variable.

In this paper, we propose to use a \emph{ranking} strategy to learn from
medical imaging data. Ranking is a type of supervised machine learning
problem that has been widely used in web search and information
retrieval~\cite{Richardson_Prakash_Brill_2006,Burges_2010} and whose goal
is to automatically construct an order from the training data.
%
We first detail how the ranking problem can be solved using binary
classifiers applied to pairs of images and then provide empirical
evidence on simulated data that our approach outperforms standard
regression techniques. Finally, we provide results on two fMRI
datasets.

\paragraph{Notations}

We write vectors in bold, $\mathbf{a} \in \RR^{n}$, matrices with capital
bold letters, $\mathbf{A} \in \RR^{n\times n}$. The dot product between
two vectors is denoted $\lb \B{a}, \B{b} \rb$. We denote by
$\|\mathbf{a}\|  = \sqrt{\lb \B{a}, \B{a} \rb}$ the $\ell_2$ norm of a
vector.



%
\section{Method: Learning to rank with a linear model}

%
A dataset consists of $n$ images (resp. volumes) containing $p$ pixels
(resp. voxels). The matrix formed by all images is denoted $\B{X} \in
\RR^{n \times p}$.

In the supervised learning setting we want to estimate a function $f$
that predicts a target variable from an image, $f: \RR^{p} \rightarrow
\mathcal{Y}$. For a classification task, $\mathcal{Y} = \{1, 2, 3,
..., k\}$ is a discrete unordered set of labels. The classification error
is then given by the number of misclassified images (0-1 loss). On the
other hand, for a regression task, $\mathcal{Y}$ is a metric space,
typically $\RR$, and the loss function can take into account the full
metric structure, \emph{e.g.} using the mean squared error.

Here, we consider a problem which shares properties of both cases. As
in the classification setting, the class labels form a finite set and
as in the regression setting there exists a natural ordering among its
elements. One option is to ignore this order and classify each data
point into one of the classes. However, this approach ignores valuable
structure in the data, which together with the high number of classes
and the limited number of images, leads in practice to poor
performance. In order to exploit the order of labels, we will use an
approach known as \emph{ranking} or \emph{ordinal regression}.

\paragraph{Ranking with binary classifiers}

Suppose now that our output space $\mathcal{Y} = \{r_1, r_2,
..., r_k\}$ verifies the ordering $r_1 \leq
r_2 \leq .. \leq r_k$ and that $f: \RR \rightarrow
\mathcal{Y}$ is our prediction function. As in~\cite{Herbrich2000}, we
introduce an increasing function $\theta: \RR \rightarrow \RR $ and a
linear function $g(\B{x}) = \lb \B{x}, \B{w} \rb$ which is related
to $f$ by
\begin{equation}
f(\B{x}) = r_i \iff g(\B{x}) \in [\theta({r}_{i-1}), \theta({r}_i)[
\end{equation} 
Given two images $(\B{x}_i, \B{x}_j)$ and their associated labels
$(y_i, y_j)$ ($y_i \neq y_j$) we form a new image
$\B{x}_i - \B{x}_j$ with label $\sign({y}_i - {y}_j)$. Because of
the linearity of $g$, predicting the correct ordering of these two
images, is equivalent to predicting the sign of $g(\B{x}_i) -
g(\B{x}_j)= \lb \B{x}_i - \B{x}_j, \B{w} \rb$~\cite{Herbrich2000}.

The learning problem is now cast into a binary classification task
that can be solved using standard supervised classification
techniques. If the classifier used in this task is a Support Vector
Machine Classifier, the model is also known as \emph{RankSVM}. One of
the possible drawbacks of this method is that it requires to consider
all possible pairs of images. This scales quadratically with the
number of training samples, and the problem soon becomes intractable
as the number of samples increases. However, specialized algorithms
exist with better asymptotic
properties~\cite{Joachims:2006:TLS:1150402.1150429}. For our study, we
used the Support Vector Machine algorithms proposed by
scikit-learn~\cite{scikit-learn}.

The main benefit of this approach is that it outputs a linear model
even when the function $\theta$ is non-linear, and thus ranking
approaches are applicable to a wider set of problems than linear
regression. Compared to multi-label classification task, where the number of
coefficient vectors increase with the number of labels (ranks), the
number of coefficients to learn in the pairwise ranking is constant,
yielding better-conditioned problems as the number of unique labels
increases.

\paragraph{Performance evaluation} Using the linear model previously introduced,
 we denote the estimated coefficients as $\B{\hat{w}} \in \RR^p$. In
 this case, the prediction function corresponds to the sign of $\lb
 \B{x}_i - \B{x}_j, \B{\hat{w}} \rb$. This means that the larger $\lb
 \B{x}_i, \B{\hat{w}} \rb$, the more likely the label associated to
 $\B{x}_i$ is to be high. Because the function $\theta$ is
 non-decreasing, one can project along the vector $\hat{\B{w}}$ to
 order a sequence of images. The function $\theta$ is generally
 unknown, so this procedure does not directly give the class
 labels. However, under special circumstances (as is the case in our
 empirical study), for example when there is a fixed number of samples
 per class this can be used to recover the target values.

Since our ranking model operates naturally on pairs of images, we will
define an evaluation measure as the mean number of label
inversions. Formally, let $(\B{x}_i, y_i)_{i=1,\dots,n}$ denote the
validation dataset and $\mathcal{P} = \{ (i,j) \textrm{ s.t. } y_i
\neq y_j \} $ the set of pairs with different labels. The prediction
accuracy is defined as the percentage of incorrect orderings for pairs
of images. When working with such a performance metric, the chance level is at 50\%:
\begin{equation}
  \label{eq:loss} \textrm{Error} = \# \left\{ (i,j) \in \mathcal{P} 
\textrm{ s.t. } (y_i - y_j) (f(\B{x}_j) - f(\B{x}_i)) \rb < 0 \right\} / \# \mathcal{P} \enspace .
\end{equation}
\paragraph{Model selection}

The Ranking SVM model has one regularization parameter denoted $C$, which
we set by nested cross-validation on a grid of 50 geometrically-spaced
values between $10^{-3}$ and $10^3$.
We use 5-folds splitting of the data: 60\% of the data is used for
training, 20\% for parameter selection and 20\% for validation.
To establish significant differences between methods we perform
20 such random data splits.

%
\section{Results}\label{section:results}
We present results on simulated fMRI data and two fMRI datasets.
\subsection{Simulation study}
\paragraph{Data generation} The simulated data $\B{X}$ contains $n=300$
volumes (size $7 \times 7 \times 7$), each one consisting of Gaussian
white noise smoothed by a Gaussian kernel with standard deviation of 2
voxels. This mimics the spatial correlation structure observed in real
fMRI data. The simulated vector of coefficients $\B{w}$ has a support
restricted to four cubic Regions of Interest (ROIs) of size ($2 \times 2
\times 2$). The values of $\B{w}$ restricted to these ROIs are $\{5,
5, -5, -5\}$.

We define the target value $\B{y} \in \RR^n$ as a logistic function of $\B{X}\B{w}$:
\begin{eqnarray}
    \B{y} = \frac{1}{1 + \exp{(-\B{X}\B{w}})} + \B{\epsilon}
    \label{eq:simulation_eq}
\end{eqnarray}
where $\B{\epsilon} \in \RR^n$ is a Gaussian noise with standard
deviation $\gamma > 0$ chosen such that the signal-to-noise ratio
verifies $\|\epsilon\| / \|\sqrt{\B{X}\B{w}}\| = 10\%$.
Finally, we split the 300 generated images
into a training set of 240 images and a validation set of other 60
images.

\paragraph{Results} We compare the ranking framework presented
previously with standard approaches. Ridge regression was chosen for
its widespread use as a regression technique applied to fMRI data. Due
to the non-linear relationship between the data and the target values,
we also selected a non-linear regression model: support vector
regression (SVR) with a Gaussian kernel~\cite{Drucker1996}. Finally,
we also considered classification models such as multi-class support
vector machines. However, due to the large number of classes and the
limited number of training samples, these methods were not competitive
against its regression counterpart and were not included in the final
comparison.

One issue when comparing different models is the qualitatively
different variables they estimate: in the regression case it is a
continuous variable whereas in the ranking settings it is a discrete
set of class labels. To make both comparable, a score function that is
applicable to both models must be used. In this case, we used as
performance measure the percentage of incorrect orderings for pairs as defined
in \eqref{eq:loss}.

\begin{figure}[bt]
\begin{center}
    \begin{minipage}{.48\linewidth}
        \includegraphics[width=\linewidth]{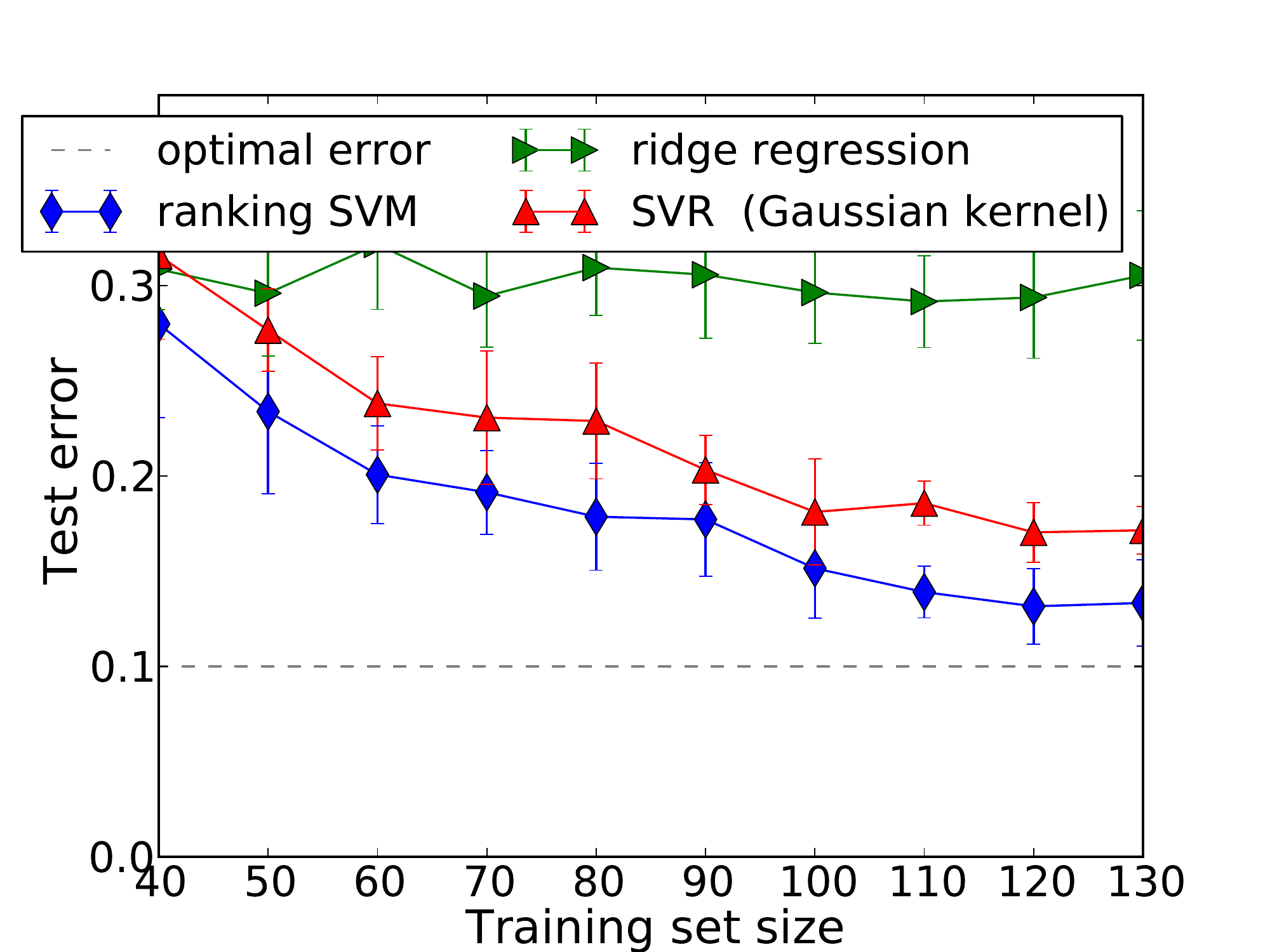}
    \end{minipage}
    \hfill
    \begin{minipage}{0.48\linewidth}
        \includegraphics[width=\linewidth]{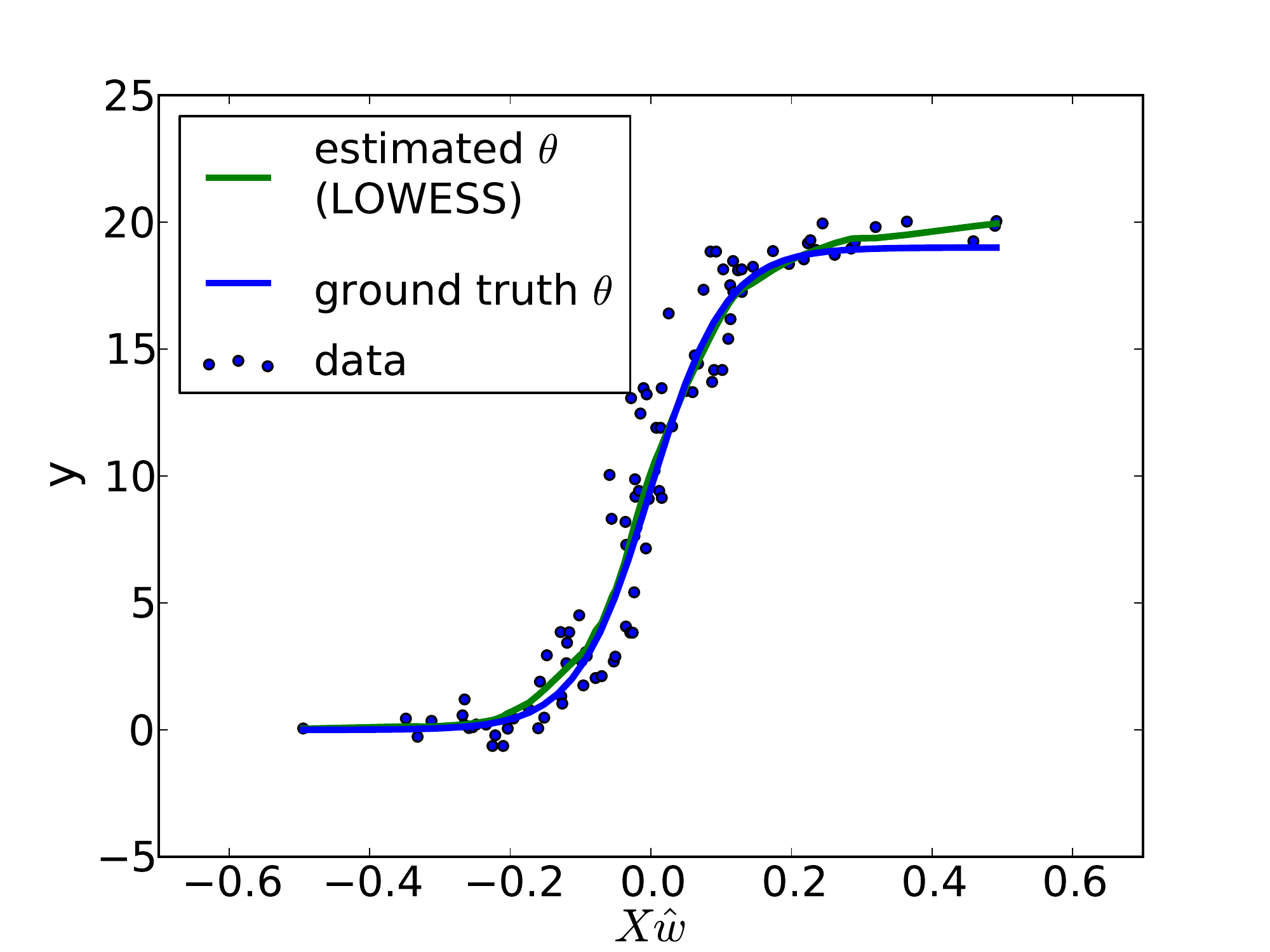}
    \end{minipage}
    \caption{\label{fig:simulation} a) Prediction accuracy as the
      training size increases. Ranking SVM performs consistently
      better than the alternative methods, converging faster to the
      empirical optimal error. b) Estimation of the $\theta$ function
      using non-parametric local regression and ranking SVM. As
      expected, we recover the logistic function introduced in the
      data generation section.}
 \end{center}
\end{figure}

Figure~\ref{fig:simulation}-a describes the performance error of the
different models mentioned earlier as a function of number of images
in the training data. We considered a validation set of 60 images and
varied the number of samples in the training set from 40 to 240. With
a black dashed line we denote the optimal prediction error, i.e. the
performance of an oracle with perfect knowledge of $\B{w}$. This error
is non zero due to noise. All of the model parameters were set by
cross-validation on the training set. The figure not only shows that
Ranking SVM converges to an optimal model (statistical consistency for
prediction), but also that it converges faster, \emph{i.e.} has lower
sample complexity than alternative approaches, thus it should
be able to detect statistical effects with less data. Gaussian kernel
SVR performs better than ridge regression and also reaches the optimal
error.  However, the underlying model of SVR is not linear and is
therefore not well-suited for interpretation
\cite{Carroll_Cecchi_Rish_Garg_Rao_2009}; moreover, it is less stable
to high-dimensional data.

As stated previously, the function $\theta$ can be estimated from the
data. In Fig.~\ref{fig:simulation}-b we use the knowledge of target
values from the validation dataset to estimate this
function. Specifically, we display class labels as a function of
$\B{X}\B{\hat{w}}$ and regularize the result using a local regression
(LOWESS). Both estimated function and ground truth overlap for most
part of the domain.

%
\subsection{Results on two functional MRI datasets}

To assess the performance of ranking strategy on real data, we
investigate two fMRI datasets. The first dataset, described
in~\cite{Cauvet2012thesis}, consists of 34 healthy volunteers scanned while
listening to 16 words sentences with five different levels of complexity. These were 1
word constituent phrases (the simplest), 2 words, 4 words, 8 words and 16
words respectively, corresponding to 5 levels of complexity which was
used as class label in our experiments. To clarify, a sentence
with 16 words using 2 words constituents is formed by a series of 8 pairs
of words. Words in each pair have a common meaning but there is meaning
between each pair. A sentence has therefore the highest complexity when all
the 16 words form a meaningful sentence.

The second dataset is described in~\cite{tom-etal:2007} and further
studied in~\cite{Jimura_Poldrack_2011} is a gambling task where each
of the 17 subjects was asked to accept or reject gambles that offered
a 50/50 chance of gaining or losing money. The magnitude of the
potential gain and loss was independently varied across 16 levels
between trials.  No outcomes of these gambles were presented during
scanning, but after the scan three gambles were selected at random and
played for real money.  Each gamble has an amount that can be used as
class label. In this experiment, we only considered gain levels,
yielding 8 different class labels. This dataset is publicly available
from \url{http://openfmri.org} as the \emph{mixed-gambles task}
dataset. The features used in both experiments are SPM $\beta$-maps,
\emph{a.k.a.} GLM regression coefficients.

Both of these datasets can be investigated with ranking, multi-label
classification or regression. We compared the approaches of regression
and ranking to test if the added flexibility of the ranking model
translates into greater statistical power. Due to the high number of
classes and limited number of samples, we found out that multi-label
classification did not perform significantly better than chance and thus
was not further considered.


To minimize the effects that are non-specific to the task we only
consider pairs of images from the same subject.



\begin{figure}[bt]
    \begin{minipage}{0.55\linewidth}
        \includegraphics[width=\linewidth]{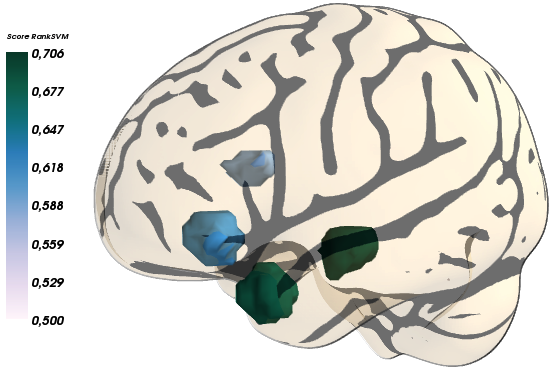}
    \end{minipage}
    \hfill
    \begin{minipage}{0.45\linewidth}
        \includegraphics[width=\linewidth]{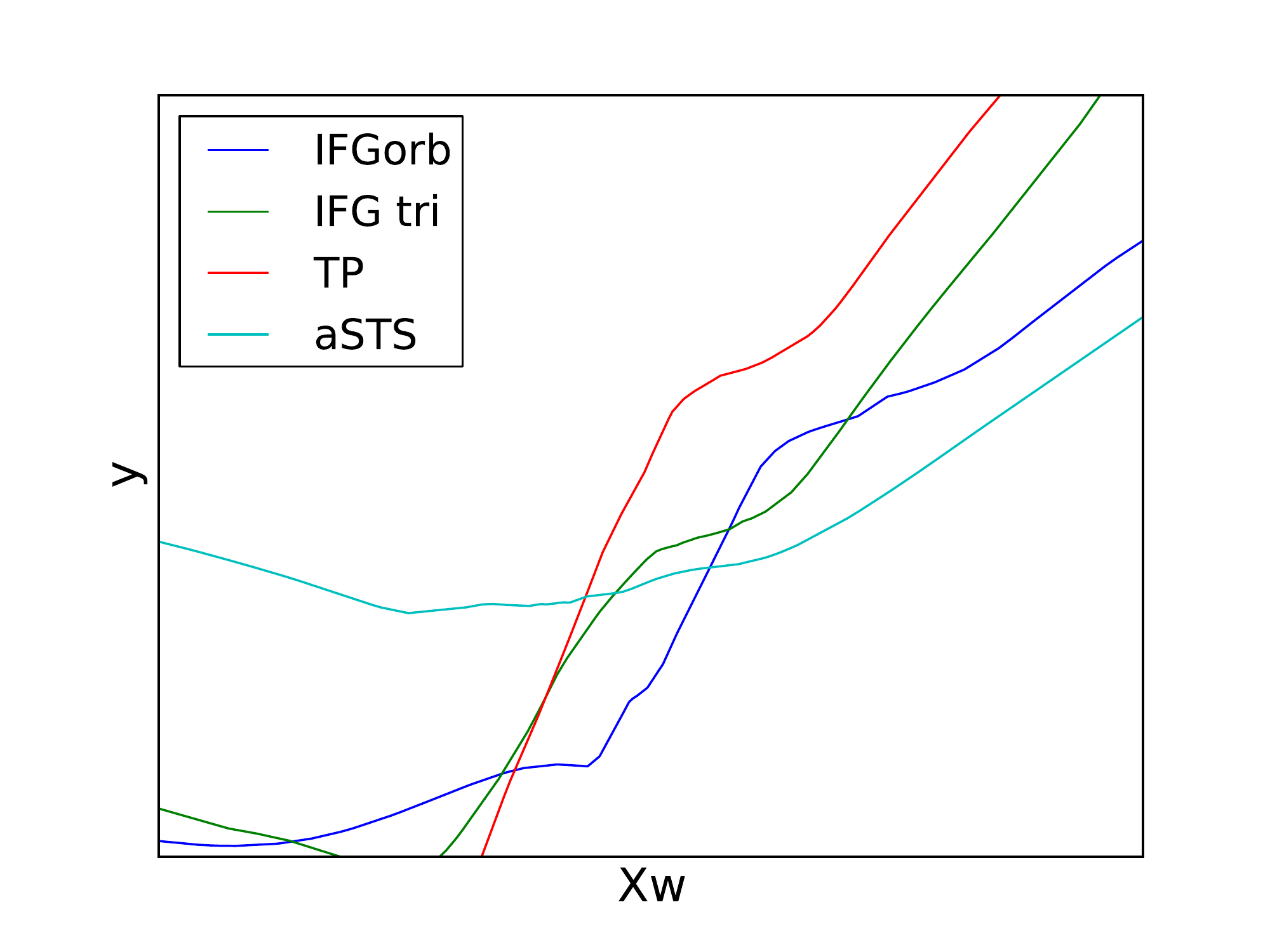}
    \end{minipage}
    \begin{minipage}{\linewidth}
        \caption{\label{fig:scores_brain}
        a) Scores obtained with the Ranking SVM on the 4 different ROIs.
        The regions with the best predictive power are the temporal pole
        the anterior superior temporal sulcus.
        b) The target variable $\B{y}$ as a function of
        $\B{X}\hat{\B{w}}$ for the four regions of interest.
        We observe that the shape of the curves
        varies across brain regions.}
    \end{minipage}
\end{figure}

The first dataset contains four manually labeled regions of interest:
Anterior Superior Temporal Sulcus (aSTS), Temporal Pole (TP), Inferior
Frontal Gyrus Orbitalis (IFGorb) and Inferior Frontal Gyrus
triangularis (IFG tri). We then compare ranking, ridge regression and
Gaussian kernel SVR models on each ROI separately. Those results
appear in the first three rows of Table~\ref{tab:performance_table}
and are denoted as language complexity with its corresponding ROI in
parenthesis. We observe that the ranking model obtains a significant
advantage on 3 of the 4 ROIs.  This could be explained by a relatively
linear effect in IFGorb or a higher noise level.The last row concerns
the second dataset, denoted gambling, where we selected the gain
experiment (8 class labels). In this case, since we were not presented
manually labeled regions of interest, we performed univariate
dimensionality reduction to 500 voxels using ANOVA before fitting the
learning models. It can be seen that the prediction accuracy are lower
compared to the first dataset. Ranking SVM however still outperforms
alternative methods.
\begin{table}[bt]
    \centering
    \begin{tabular}{|l|c|c|c|c|}
        \hline
        & RankSVM & Ridge & SVR & P-val\\
        [0.5ex]
        \hline
        lang. comp (aSTS)&{\bf 0.706} & 0.661 & 0.625 & 2e-3***\\
        lang. comp. (TP)&{\bf 0.687} & 0.645 & 0.618 & 7e-4***\\
        lang. comp. (IFGorb)&{\bf 0.619} & 0.609 & 0.539 & 0.3\\
        lang. comp. (IFG tri)&{\bf 0.585} & 0.566 & 0.533 & 5e-2*\\
        gambling & {\bf 0.58} & 0.56 & 0.53 & 1e-2**\\
        \hline
    \end{tabular}
    \caption{
    \label{tab:performance_table}
    Prediction accuracy of the ranking strategy on two real fMRI datasets:
    language complexity (lang. comp.) in 3 ROIs and gambles.
    As a comparison, scores obtained with
    alternative regression techniques (ridge regression and SVR using a non-linear
    Gaussian kernel) are presented. As confirmed by a Wilcoxon paired
    test between errors obtained for each fold using Ranking SVM and ridge regression,
    Ranking SVM leads to significantly better scores than other approaches
    on 4 of the 5 experiments.}
\end{table}
Locations in the brain of the ROIs for the first dataset, with a color
coding for their predictive power are presented in
Fig.~\ref{fig:scores_brain}-a.
As shown previously in the simulated dataset,
Fig.~\ref{fig:scores_brain}-b shows the validation data projected
along the coefficients of the linear model and regularized using
LOWESS local regression for each one of the four highest ranked regions
of interest. Results show that the link function between
$\B{X}\hat{\B{w}}$ and the target variable $\B{y}$ (denoted $\theta$
in the methods section) varies in shape across ROIs, suggesting that
the BOLD response is not unique over the brain.
%




%
\section{Discussion and conclusion}

In this paper, we describe a ranking strategy that addresses a common
use case in the statistical analysis of medical images, which is the
prediction of an ordered target variable. Our contribution is to
formulate the variable quantification problem as a ranking problem. We
present a formulation of the ranking problem that transforms the task
into a binary classification problem over pairs of images. This
approach makes it possible to use efficient linear classifiers while
coping with non-linearities in the data. By doing so we retain the
interpretability and favorable behavior in high-dimension of linear
models.

From a statistical standpoint, mining medical images is challenging
due to the high dimensionality of the data, often thousands of
variables, while the number of images available for training is small,
typically a few hundreds. In this regard, the benefit of our approach
is to retain a linear model with few parameters. It is thus better
suited to medical images than multi-class classification.

On simulations we have shown that our problem formulation leads to a
better prediction accuracy and lower sample complexity than
alternative approaches such as regression or classification
techniques. We apply this method to two fMRI datasets and discuss
practical considerations when dealing with fMRI data. We confirm the
superior prediction accuracy compared to standard regression techniques.



%

%
%
\bibliography{biblio}{}

\end{document}